%% file: main.tex
\documentclass[11pt]{article}
\usepackage{acl2015}
\usepackage{times}
\usepackage{url}
\usepackage{latexsym}
\usepackage{tikz}
\usepackage{graphicx}
\usepackage{amsmath}
\usepackage{mathtools}
\usepackage{amssymb}

\title{Part-of-Speech Tagging with Bidirectional Long Short-Term Memory \\ Recurrent Neural Network }

\author{Peilu Wang$^{1,2}$, Yao Qian$^3$, Frank K. Soong$^2$, Lei He$^2$, Hai Zhao$^1$\\
  $^1$Shanghai Jiao Tong University, Shanghai, China\\
  $^2$Microsoft Research Asian, Beijing, China \\
  $^3$Educational Testing Service Research, USA\\
  {\tt \{v-peiwan, frankkps, helei\}@microsoft.com,}\\
  {\tt zhaohai@cs.sjtu.edu.cn, yqian@ets.org} }

\date{}

\begin{document}
\maketitle
\begin{abstract}
Bidirectional Long Short-Term Memory Recurrent Neural Network (BLSTM-RNN) has been shown to be very effective for tagging sequential data, e.g. speech utterances or handwritten documents.
While word embedding has been demoed as a powerful representation for characterizing the statistical properties of natural language.
In this study, we propose to use BLSTM-RNN with word embedding for part-of-speech (POS) tagging task.
When tested on Penn Treebank WSJ test set, a state-of-the-art performance of 97.40 tagging accuracy is achieved.
Without using morphological features, this approach can also achieve a good performance comparable with the Stanford POS tagger.

\end{abstract}

\input{part.intro}

\input{part.methods}

\input{part.exp}

\section{Conclusions}
In this paper, BLSTM RNN is proposed for POS tagging and training word embedding.
Combined with word embedding trained on big unlabeled data, this approach gets state-of-the-art accuracy on WSJ test set without using rich morphological features.
BLSTM RNN with word embedding is expected as an effective solution for tagging tasks and worth further exploration.
\bibliographystyle{acl}
\bibliography{acl2015}

\end{document}

%% file: part.intro.tex
\section{Introduction}

Bidirectional long short-term memory \cite{1997_Sepp_NC_LongShort,1997_Mike_SP_BidirectionalRecurrent} (BLSTM) is a type of recurrent neural network (RNN) that can incorporate contextual information from long period of fore-and-aft inputs. 
It has been proven a powerful model for sequential labeling tasks.
For applications in natural language processing (NLP), it has helped achieve superior performance in language modeling \cite{2012_Martin_INTERSPEECH_LSTMNeural,2015_Martin_ASLP_FromFeedforward}, language understanding \cite{2013_Kaisheng_INTERSPEECH_RecurrentNeural}, and machine translation \cite{2014_Martin_EMNLP_TranslationModeling}.
Since part-of-speech (POS) tagging is a typical sequential labeling task, it seems natural to expect BLSTM RNN can also be effective for this task.

As a neural network model, it is awkward for BLSTM RNN to make use of conventional NLP features, such as morphological features.
Since these features are discrete and has to be represented as one-hot vector to be used, using rich this type of features leads to too large input layer to maintain and update.
Therefore, we avoid using such features except word form and simple capital features, instead we involve word embedding.
Word embedding is a low dimensional real-valued vector used to represent word.
It is considered containing part of syntactic and semantic information and has shown a very attractive feature for various of language processing tasks \cite{2008_Collobert_ICML_AUnified,2010_Joseph_ACL_WordRepresentations,2011_Ronan_JMLR_NaturalLanguage}. 
Word embedding can be obtained by training a neural network model, especially, a neural network language model \cite{2006_Bengio_IML_NeuralProbabilistic,2010_Mikolov_INTERSPEECH_RecurrenntNeural} or a neural network designed for a specific task \cite{2011_Ronan_JMLR_NaturalLanguage,2013_Mikolov_ARXIV_EfficientEstimation,2014_Jeffrey_EMNLP_GloveGlobal}.
Currently many word embeddings trained on quite large corpora are available on line.
However, these embeddings are trained by neural networks that are very different from BLSTM RNN. 
This inconsistency is supposed as an shortcoming to make the most of these trained word embeddings. 
To conquer this shortcoming, we also propose a novel method to train word embedding on unlabeled data with BLSTM RNN.

The main contributions of this work include: 
First, it shows an effective way to use BLSTM RNN for POS tagging task and achieves a state-of-the-art tagging accuracy.
Second, a novel method for training word embedding is proposed.
Finally, we demonstrate that competitive tagging accuracy can be obtained without using morphological features, which makes this approach more practical to tag a language that lacks of necessary morphological knowledge.

%% file: part.methods.tex
\section{Methods}

\subsection{BLSTM RNN for POS Tagging}

\begin{figure}[h]
\small
\centering
\input{graph.rnntagging.tex}
\caption{BLSTM RNN for POS tagging}\label{rnntagging}
\end{figure}
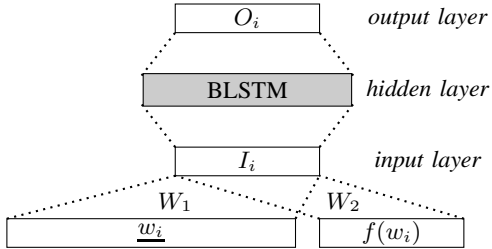

Given a sentence $w_1, w_2, ..., w_n$ with tags $y_1, y_2, ..., y_n$, BLSTM RNN is used to predict the tag probability distribution of each word.
The usage is illustrated in Figure \ref{rnntagging}.
Here $\underline{w_i}$ is the one hot representation of the current word.
It is a binary vector of dimension $|V|$ where $V$ is the vocabulary.
To reduce $|V|$, each letter in input word is transferred into lower case.
To still keep the upper case information, a function $f(w_i)$ is introduced to indicate the original case information of word $w_i$.
More specifically, $f(w_i)$ returns a three-dimensional binary vector to tell if $w_i$ is full lowercase, full uppercase or leading with a capital letter.
The input vector $I_i$ of the neural network is computed as:
\begin{equation*}
I_i=W_1 \underline{w_i}+W_2 f(w_i)
\end{equation*}
where $W_1$ and $W_2$ are weight matrixes connecting two layers.
$W_1\underline{w_i}$ is the word embedding of $w_i$ which has a much smaller dimension than $\underline{w_i}$.
In practice, $W_1$ is implemented as a lookup table, $W_1\underline{w_i}$ is returned by referring to the word embedding of $w_i$ stored in this table.
To use word embeddings trained by other task or method, we just need to initialize this lookup table with those external embeddings.
For words without corresponding external embeddings, their word embeddings are initialized with uniformly distributed random values, ranging from -0.1 to 0.1.
The implementation of BLSTM layer is detailed descripted in \cite{2012_Alex_BOOK_SupervisedSequence} and therefore is skipped in this paper.
This layer incorporates information from the past and future histories when making prediction for current word and is updated as a function of the entire input sentence.
The output layer is a softmax layer whose dimension is the number of tag types.
It outputs the tag probability distribution of input word $w_i$.
All weights are trained using backpropagation and gradient descent algorithm to maximize the likelihood on training data:
\begin{equation*}
\prod_{i\in{1,...,n}} P_i(y_i|w_1, w_2, ..., w_n)
\end{equation*}
The obtained probability distribution of each step is supposed independent with each other.
The utilization of contextual information strictly comes from the BLSTM layer.
Thus, in inference phase, the likeliest tag $y'_i$ of input word $w_i$ can just be chose as:
\begin{equation*}
y'_i= arg\max\limits_{t\in {1,...,m}}  P_i(t|w_1, w_2, ..., w_n)
\end{equation*}
where $m$ is the number of tag types.

\subsection{Word Embedding}
In this section, we propose a novel method to train word embedding on unlabeled data with BLSTM RNN.
In this approach, BLSTM RNN is also used to do a tagging task, but only has two types of tags to predict: incorrect/correct.
The input is a sequence of words which is a normal sentence with some words replaced by randomly chosen words.
For those replaced words, their tags are 0 (incorrect) and for those that are not replaced, their tags are 1 (correct).
Although it is possible that some replaced words are also reasonable in the sentence, they are still considered ``incorrect''. 
Then BLSTM RNN is trained to minimize the binary classification error on the training corpus.
The neural network structure is the same as that in Figure \ref{rnntagging}.
When the neural network is trained, $W_1$ contains all trained word embeddings.

%% file: graph.rnntagging.tex
\definecolor{royalblue}{RGB}{67,107,149}
\tikzset{
wordvector/.style={
  rectangle,
  draw,
  text width=12em,
  text centered,
  minimum height=1.2em,
  inner sep=0em
},
input/.style={
  rectangle,
  draw,
  text width=6em,
  text centered,
  minimum height=1.2em,
  inner sep=0em
},
output/.style={
  rectangle,
  draw,
  text width=6em,
  text centered,
  minimum height=1.2em,
  inner sep=0em
},
wordembedding/.style={
  rectangle,
  draw,
  text width=6em,
  text centered,
  minimum height=1.2em,
  inner sep=0em
},
NN/.style={
  rectangle,
  draw,
  text width=8em,
  text centered,
  minimum height=1.2em,
  fill=black!20,
},
block_noborder/.style={
  rectangle,
  draw=none,
  text width=8em,
  text centered,
  minimum height=1em,
  inner sep=0em,
  font={\itshape}
},
weightsymbol/.style={
  rectangle,
  draw=none,
  text width=2em,
  text centered,
  minimum height=1em,
  inner sep=0em,
},
}

\begin{tikzpicture}[x=1em, y=1em, >=stealth]

\node[wordvector] (w1) at (0,0) {$\underline{w_{i}}$};
\node[input] (w2) at (10,0){$f(w_{i})$};

\node[wordembedding] (we1) at (4,3) {$I_i$};
\node[block_noborder](tinput) at (11.5,3) {input layer};

\node[NN] (NN) at (4,6) {BLSTM};
\node[block_noborder](tinput) at (11.5,6) {hidden layer};

\node[output] (w3) at (4,9){$O_i$};
\node[block_noborder](tinput) at (11.5,9) {output layer};

\draw [-,dotted,thick] (-6,0.6) -- (1,2.4);
\draw [-,dotted,thick] (6,0.6) -- (7,2.4);
\node[weightsymbol](w1) at (1,1.3) {$W_1$};

\draw [-,dotted,thick] (7,0.6) -- (1,2.4);
\draw [-,dotted,thick] (13,0.6) -- (7,2.4);
\node[weightsymbol](w2) at (8,1.3) {$W_2$};

\draw [-,dotted,thick] (1,3.6) -- (-0.4,5.4);
\draw [-,dotted,thick] (7,3.6) -- (8.4,5.4);

\draw [-,dotted,thick] (-0.4,6.6) -- (1,8.4);
\draw [-,dotted,thick] (8.4,6.6) -- (7,8.4);

\end{tikzpicture}

%% file: part.exp.tex
\section{Experiments}

BLSTM RNN systems in our experiments are implemented with CURRENT \cite{2014_Felix_JMLR_IntroducingCurrennt}, a machine learning library for RNN which adopts GPU acceleration.
The activation function of input layer is identity function, hidden layer is logistic function, while the output layer uses softmax function for multiclass classification. 
Neural network is trained using statistical gradient descent algorithm with constant learning rate.

\subsection{Corpora}

The part-of-speech tagged data used in our experiments is the Wall Street Journal data from Penn Treebank III \cite{1993_Mitchell_CL_BuildingaLarge}.
Training, development and test sets are split following setup in \cite{2002_Michael_ACL_DiscriminativeTraining}.
Table \ref{splitcorpus} lists the detailed information of the three data sets.

\begin{table}[h]
\centering
\small
\begin{tabular}{|c|c|c|c|}
\hline
\textbf{Data Set} & \textbf{Sections} & \textbf{Sentences} & \textbf{Tokens} \\
\hline
Training & 0-18 & 38,219 & 912,344 \\
\hline
Develop & 19-21 & 5,527 & 131,768 \\
\hline
Test & 22-24 & 5,462 & 129,654 \\
\hline
\end{tabular}
\caption{Splits of WSJ corpus}\label{splitcorpus}
\vskip -1.6ex
\end{table}

To train word embedding, we uses North American news \cite{2008_David_LDC_NorthAmerican} as the unlabeled data.
This corpus contains about 536 million words.
It is tokenized using the Penn Treebank tokenizer script \footnote{\url{https://www.cis.upenn.edu/~treebank/tokenization.html}}.
All consecutive digits occurring within a word are replaced with the symbol ``\#''.
For example, both words ``Tel192'' and ``Tel6'' are transferred to the same word ``Tel\#''.

\subsection{Hidden Layer Size}

\begin{figure}[ht]
\includegraphics[width=3in]{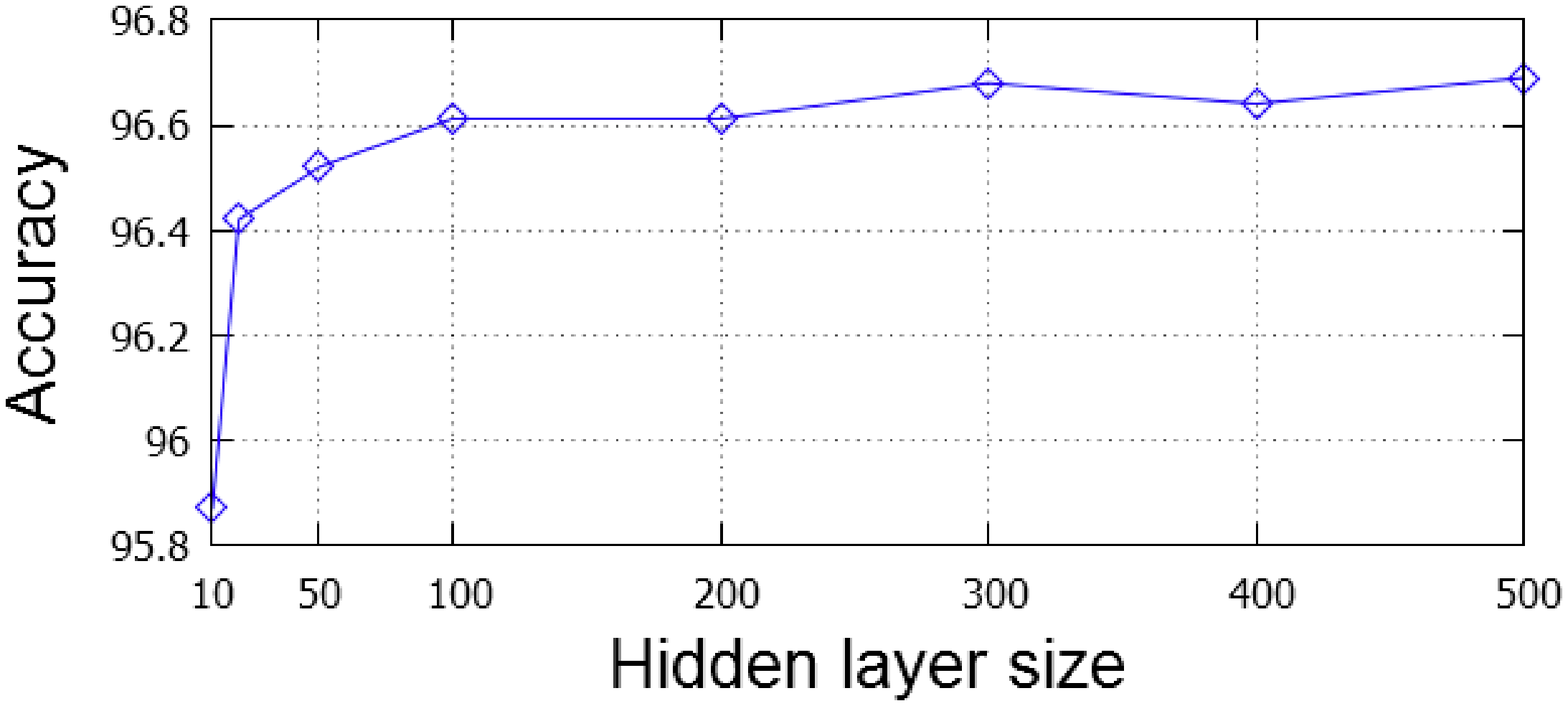}
\vskip -1.2ex
\caption{Accuracy of different hidden layer sizes}\label{explorehidden}
\vskip -1.6ex
\end{figure}

We evaluate different sizes of hidden layer in BLSTM RNN to pick up the best structure for later experiments.
The input layer size is set to 100 and output layer size is fixed as 45 in all experiments.
The accuracies on WSJ test set are shown in Figure \ref{explorehidden}.
It shows that hidden layer size has a limited impact on performance when it becomes large enough.
To keep a good trade-off of accuracy, model size and running time, we choose 100 which is the smallest layer size to get ``reasonable'' performance as the hidden layer size in all the following experiments.

\subsection{POS Tagging Accuracies}

Table \ref{postagging} compares the performance of our systems with other baseline systems.

\begin{table}[h]
\centering
\small
\begin{tabular}{|l|c|}
\hline
\textbf{Sys} & \textbf{Acc (\%)} \\
\hline
\cite{2003_Toutanova_NAACL_FeatureRich} & 97.24 \\
\hline
\cite{2012_Liang_NAACLHLT_StructuredPerceptron} & 97.35 \\
\hline
\cite{2011_Ronan_JMLR_NaturalLanguage} NN & 96.36 \\
\hline
\cite{2011_Ronan_JMLR_NaturalLanguage} NN+WE & 97.20 \\
\hline
\hline
\textbf{BLSTM-RNN} & 96.61 \\
\hline
\hline
\textbf{BLSTM-RNN+WE(10m)} & 96.61 \\
\hline
\textbf{BLSTM-RNN+WE(100m)} & 97.10 \\
\hline
\textbf{BLSTM-RNN+WE(all)} & 97.26 \\
\hline
\hline
\textbf{BLSTM-RNN+WE(all)+suffix2} & \textbf{97.40} \\
\hline
\end{tabular}
\caption{POS tagging accuracies on WSJ test set.}\label{postagging}
\vskip -1.6ex
\end{table}

\textbf{Baseline systems}. 
Four typical systems are chosen as baseline systems.
\cite{2003_Toutanova_NAACL_FeatureRich} is one of the most commonly used approaches which is also known as Stanford tagger.
\cite{2012_Liang_NAACLHLT_StructuredPerceptron} is the system reports best accuracy on WSJ test set (97.35\%).
In fact, \cite{2009_Drahom_EACL_SemiSupervised} reports a higher accuracy (97.44\%), but this work relies on multiple trained taggers and combines their tagging results.
Here we focus on single model tagging algorithm and therefore do not include this work as baseline.
Besides, \cite{2014_Robert_COLING_FastHigh} (97.34\%) and \cite{2007_Libin_ACL_GuidedLearning} (97.33\%) also reach accuracy above 97.3\%.
These two systems plus \cite{2012_Liang_NAACLHLT_StructuredPerceptron} are considered as current state-of-the-art systems.
All these systems rely on rich morphological features.
In contrast, \cite{2011_Ronan_JMLR_NaturalLanguage} NN only uses word form and capital features. 
\cite{2011_Ronan_JMLR_NaturalLanguage} NN+WE also incorporates word embeddings trained on unlabeled data like our approach.
The main difference is that \cite{2011_Ronan_JMLR_NaturalLanguage} uses feedforward neural network instead of BLSTM RNN.

\textbf{BLSTM-RNN} is the system described in Section 2.1 which only uses word form and capital features.
The vocabulary we used in this experiment is all words appearing in WSJ Penn Treebank training set, merging with the most common 100,000 words in North American news corpus, plus one single ``UNK'' symbol for replacing all out of vocabulary words.

Without the help of morphological features, it is not surprising that \textbf{BLSTM-RNN} falls behind the state-of-the-art system.
However, \textbf{BLSTM-RNN} surpasses \cite{2011_Ronan_JMLR_NaturalLanguage} NN which is also neural network based method and uses the same input features.
It is consistent with \cite{2014_Raul_INTERSPEECH_ProsodyContour,2014_Yuchen_INTERSPEECH_TTSSythesis}, in which BLSTM RNN outperforms feedforward neural network.

\begin{table*}[t]
\centering
\small
\begin{tabular}{|c|c|c|p{13em}|c|c|c|}
\hline
\textbf{WE} & \textbf{Dim} & \textbf{Vocab Size} & \textbf{Train Corpus (Toks \#)} & \textbf{OOV} & \textbf{Acc (\%)} \\
\hline
\cite{URL_RNNLM} & 80 & 82K & Broadcast news (400M) & 0.31 & 96.91 \\
\hline
\cite{URL_WordRepresent} & 100 & 269K & RCV1 (37M) & 0.18 & 96.81 \\
\hline
\cite{URL_SENNA} & 50 & 130K & RCV1+Wiki (221M+631M)  & 0.22 & 97.02 \\
\hline
\cite{URL_Word2Vec} & 300 & 3M & Google news (10B)  & 0.17 & 96.86 \\
\hline
\cite{URL_Glove}1 & 100 & 400K & Wiki (6B)  & 0.13 & 97.12 \\
\hline
\cite{URL_Glove}2 & 100 & 1193K & Twitter (27B)  & 0.25 & 97.00 \\
\hline
\hline
BLSTM RNN WE & 100 & 100K & North American news (536M) & 0.17 & 97.26 \\
\hline
\end{tabular}
\caption{Comparison of different word embeddings.}\label{wecmp}
\vskip -1.6ex
\end{table*}

\textbf{BLSTM-RNN+WE}. 
To construct corpus for training word embeddings, about 20\% words in normal sentences of North American news corpus are replaced with randomly selected words.
Then BLSTM RNN is trained to judge which word has been replaced as described in Section 2.2.
The vocabulary for this task contains the 100,000 most common words in North American news corpus and one special ``UNK'' symbol.
When training is finished, word embedding lookup table ($W_1$) in BLSTM RNN for POS tagging is initialized with the trained word embeddings.
The following training and testing are the same as previous experiment.

Table \ref{postagging} shows the results of using word embeddings trained on the first 10 million words (\textbf{WE(10m)}), first 100 million words (\textbf{WE(100m)}) and all 530 million words (\textbf{WE(all)}) of North American news corpus.
While \textbf{WE(10m)} does not show much help for the improvement, \textbf{WE(100m)} and \textbf{WE(all)} significantly boosts the performance.
It shows that BLSTM RNN can benefit from word embeddings trained on large unlabeled corpus and larger training corpus leads to a better performance.
This suggests that the result may be further improved by using even bigger unlabeled data set.
With the help of GPU, \textbf{WE(all)} can be trained in about one day (23 hrs).
The training time increases linearly with the training corpus size.

\textbf{WE(all)} reduces over 20\% error rate of  \textbf{BLSTM-RNN} and lets the result comparable with \cite{2003_Toutanova_NAACL_FeatureRich}.
Note that this result is obtained without using any morphological features.
Current state-of-the-art systems \cite{2014_Robert_COLING_FastHigh,2007_Libin_ACL_GuidedLearning,2012_Liang_NAACLHLT_StructuredPerceptron} all utilize morphological features proposed in \cite{1996_Adwait_EMNLP_AMaximum} which involves $n$-gram prefix and suffix ($n$ = 1 to 4).
Moreover, \cite{2007_Libin_ACL_GuidedLearning} also involves prefix and suffix of length from 5 to 9.
\cite{2014_Robert_COLING_FastHigh} adds extra elaborately designed features, including flags indicating if word ends with $-ed$ or $-ing$, etc.
In practice, many languages with rich morphological forms lack of necessary or effective morphological processing tools. 
In these cases, a POS tagger that does not rely on morphological features is more realistic for use.

\textbf{BLSTM-RNN+WE(all)+suffix2}. 
In this experiment, we add bigram suffix of each word as extra feature.
These last 2 characters are represented as one-hot vector and appended to the original extra feature vector ($f(w_i)$).
The other configuration follows \textbf{BLSTM-RNN+WE(all)}.
The additional feature furthermore pushes up the accuracy and lets the approach get the state-of-the-art performance (\textbf{97.40\%}).
However, adding more morphological features such as trigram suffix does not further improve the performance.
One possible reason is that adding such feature brings a much longer extra feature vector which needs retuning parameters such as learning rate and hidden layer size to get the optimum performance.


\subsection{Different Word Embeddings}
In this experiment, six types of published well-trained word embeddings are evaluated.
The basic information of involved word embeddings and results are listed in Table \ref{wecmp} where RCV1 represents the Reuters Corpus Volume 1 news set.
The OOV (out of vocabulary) column indicates the rate of words in vocabulary of BLSTM RNN for POS tagging that are not covered by external word embedding vocabulary. 
The usage of word embeddings is the same as in \textbf{BLSTM-RNN+WE} experiment except that input layer size here is equal to the dimension of external word embedding.

All word embeddings bring about higher accuracy.
However, none of them can enhance BLSTM RNN tagging to get a competitive accuracy, despite of larger corpora that they are trained on and lower OOV rate.
\cite{URL_Glove}1 (97.12\%) has the highest accuracy among them but it is still lower than \cite{2003_Toutanova_NAACL_FeatureRich} (97.24\%).
Although more experiments are needed to judge which word embeddings are better, this experiment at least shows word embeddings trained by BLSTM RNN are essential in our POS tagging approach to achieve a superior performance.

